\documentclass{article}
\usepackage{spconf,amsmath,graphicx}

\usepackage[ruled,vlined]{algorithm2e}
\usepackage{booktabs}
\usepackage[flushleft]{threeparttable}
\SetKwComment{Comment}{$\triangleright$\ }{}

\title{Model-Based Approach for Measuring the Fairness in ASR}

\name{Zhe Liu, Irina-Elena Veliche, Fuchun Peng}
\address{Facebook AI, Menlo Park, CA, USA}

\begin{document}
\ninept
\maketitle

\begin{abstract}
The issue of fairness arises when the automatic speech recognition (ASR) systems do not perform equally well for all subgroups of the population. In any fairness measurement studies for ASR, the open questions of how to  control the nuisance factors, how to handle unobserved heterogeneity across speakers, and how to trace the source of any word error rate (WER) gap among different subgroups are especially important - if not appropriately accounted for, incorrect conclusions will be drawn. In this paper, we introduce mixed-effects Poisson regression to better measure and interpret any WER difference among subgroups of interest. Particularly, the presented method can effectively address the three problems raised above and is very flexible to use in practical disparity analyses. We demonstrate the validity of proposed model-based approach on both synthetic and real-world speech data.
\end{abstract}


\section{Introduction}
Automatic speech recognition (ASR) systems are getting better with the advent of new technologies, however, the issue of \emph{fairness} arises when these tools do not perform equally well for all subgroups of the population \cite{tatman2017effects, tatman2017gender, mehrabi2019survey, koenecke2020racial, martin2020understanding}. The concern of fairness, is not limited to speech recognition tasks, but also comes to light in other machine learning applications, including facial recognition \cite{garvie2016facial, xu2020investigating}, natural language processing \cite{blodgett2016demographic, caliskan2017semantics}, and healthcare \cite{obermeyer2019dissecting}.

The fairness issue was highlighted most recently by authors in \cite{koenecke2020racial} who found that five state-of-the-art ASR systems showed substantial racial disparities, having an average word error rate (WER) of 0.35 for black speakers compared with 0.19 for white speakers. Here, WER serves as the most widely used metric for measuring the performance of an ASR system, which is derived from the Levenshtein distance \cite{navarro2001guided} working at the word level:
\begin{equation}
    \text{WER}=\frac{\sum_{s=1}^n e_s}{\sum_{s=1}^n m_s}
\end{equation}
where $m_s$ is the number of words in the $s$th sentence (i.e. reference text of audio) of the evaluation dataset, and $e_s$ represents the sum of insertion, deletion, and substitution errors computed from the dynamic string alignment of the recognized word sequence with the reference word sequence.

In most of previous research studies on measuring fairness in ASR, WER was computed per each subgroup (e.g. black speakers versus white speakers) and from the comparison of these WER numbers, conclusions were drawn on whether significant disparities exist among these subgroups of interest. Although this is a very simple way for fairness measurement, there are several open questions that have not yet been properly addressed in these analyses.

First, how to effectively control the \emph{nuisance} factors which may affect the measured results but are not of primary interest? For example, we need to deal with any unbalanced gender or age distribution of speakers in a racial disparity study; otherwise, it would be difficult to tell whether any WER gap among different racial groups is due to the race factor or any nuisance factor of gender or age. Propensity-score matching was utilized in \cite{koenecke2020racial} to select a subset of audio snippets of white and black speakers with similar distributions of age and gender. However, this matching procedure suffers from discarding ``unmatched'' but informative samples from the analysis and in some scenarios, a good matching might not even exist.

Secondly, how to appropriately take into account speaker-level effect on measured WERs and handle unobserved \emph{heterogeneity} across different speakers? Speech recognition accuracy on utterances from the same speaker could be highly correlated and a neglect of such dependency structure in speech data might lead to underestimated variance and wrong conclusions. Moreover, it is typically reasonable to consider speakers were being randomly selected from the subgroups of interest, and we are not particularly interested in these specific speakers, but in the population that they represent.

Third, how to efficiently trace the source of any WER gap among different subgroups, that is, does such disparity mainly come from phonetic, phonological, prosodic characteristics, or grammatical, lexical, semantic characteristics, or both? Perplexity was used in existing literature to evaluate the grammatical, lexical, semantic properties of disparities across different subgroups. However, more advanced approaches are still called for to provide deeper insights and analyses on whether language model or acoustic model should account for the overall disparities in WERs if at all.

In this paper, we present a model-based approach to better measure the fairness issue in ASR and study any performance disparities across different subgroups of our interest. In particular, we introduce \emph{mixed-effects Poisson regression} \cite{mccullagh1989generalized, baltagi2008econometric, faraway2016extending}, treating utterance-level word errors as the regression response, logarithm number of words in the reference text as an offset, speaker identification as a random effect, subgroup label of interest and any other explanatory or confounding variables as fixed effects. The presented method can address the three problems that we previously raised, and is very flexible to use. As classical and powerful statistical tools, mixed-effects model and Poisson regression are not new in analyzing real-world scientific problems. But to the best of our knowledge, our work is the first to introduce sophisticated statistical regression-based approach to investigate fairness issues in ASR and illustrate how it helps measure and interpret any WER difference across different subgroups of the population in any disparity study. In particular, our proposed method prevents underestimating the standard errors and avoids drawing false positive conclusions on non-fairness.

The rest of this paper is structured as follows. Section \ref{methodology} introduces the use of mixed-effects Poisson regression on ASR fairness. Sections \ref{simulation} and \ref{real} demonstrate the validity of the proposed method on synthetic and real-world speech data. We conclude in Section \ref{conclusion}.

\section{Methods}
\label{methodology}
In this section, we present mixed-effects Poisson regression method and illustrate how it helps measure any WER gap between different subgroups in disparity studies.

Suppose we would want to investigate the fairness in ASR with respect to some factor variable of primary interest (e.g.~gender of speakers). For the $s$th utterance in the evaluation dataset, we denote its factor level as $f(s)$ (e.g. male speaker or female speaker), where $f$ is a deterministic function with $l$ as the total number of levels. We aim to test whether the effect of this factor is statistically significant on measured WER results across its different groups of levels.

\subsection{Poisson Regression for Measuring Fairness}
Poisson regression serves as an appropriate approach to model rate data \cite{cameron2013regression}, where the rate is a count of events (e.g.~word errors in our use case) divided by some measure of that unit's exposure (e.g.~number of words in the reference). An \emph{offset} variable is needed to scale the modeling of the mean parameter in Poisson regression with a log link. Here, the underlying assumption is that the number of word errors occurred in any utterance is proportional to the number of words in the corresponding reference text.

More specifically, to measure the effect of factor $f(\cdot)$ on WER results across $l$ different subgroups, the vanilla Poisson regression model is described as follows:
\begin{align}
\label{iid}
C_{s} & \overset{\text{i.i.d.}}{\sim} Poisson(\lambda_s) \\
\label{model}
\log(\lambda_s) &= \log(N_s) + \mu_{f(s)}
\end{align}
where $C_s$ is the count of word errors (sum of insertion, deletion, and substitution errors), $\lambda_s$ is the Poisson (mean) parameter, $N_s$ is the number of words in the reference text for the $s$th utterance in the evaluation dataset, and $\mu_{f(s)}$ refers to the factor effect corresponding to the subgroup of $f(s)$. The notation of i.i.d in  (\ref{iid}) represents independent and identically distributed, where we will revisit this distribution assumption later in this section. Note that any utterance with empty reference text should be removed from the analysis since it does not provide any insight on fairness measurement.

This model can be fitted using maximum likelihood approach. Standard statistical testing, for example, likelihood ratio test (LRT) \cite{king1989unifying}, can be conducted afterwards to compute the $p$-value of the factor $f(\cdot)$ on measured WER results.

Sometimes, it is possible to analyze rate data using a binomial response model. However, in our application, number of word errors occurred in some utterance could be larger than the total number of words in the reference, which limits the use of binomial regression here. If the rate is relatively small, the Poisson approximation to the binomial is effective.

One of the key features of Poisson distribution is that the variance equals the mean. In certain circumstances, it is found that the empirical variance is greater than the mean, known as \emph{overdispersion} \cite{berk2008overdispersion, ver2007quasi}. A common reason is the omission of relevant explanatory variables, or the present of dependent samples, which we will explore more in the next two subsections.

\subsection{Poisson Regression with Explanatory Covariates}
It is natural and flexible to extend the vanilla Poisson regression model (\ref{iid})~(\ref{model}) to include additional explanatory or confounding covariates, which can be utilized to capture effects of nuisance variables on WERs among different subgroups:
\begin{align}
\label{model_v2}
\log(\lambda_s) &= \log(N_s) + \mu_{f(s)} + \theta^T x_s
\end{align}
Here, $x_s$ represents the vector of any explanatory variables in the regression model and $\theta$ refers to the coefficient parameter vector that shall be learned. For example, in a racial disparity analysis, we would want to add the gender or age information of speakers to the regression model in order to control any nuisance effects.

In particular, we can include any representative vector \cite{mikolov2013efficient}, for example, sentence embedding, of the true reference text per each utterance as extra explanatory variables, which would help us understand the source of any performance gap between different subgroups of interest. For instance, after controlling the effect of sentence embedding covariates that account for grammatical, lexical, or semantic characteristics, if the factor effect of interest is still statistically significant, we can tell that phonetic, phonological, or prosodic characteristics substantially contribute to the overall disparities among different subgroups of the factor $f(\cdot)$. Thus this can provide insights on whether language model or acoustic model should be responsible for the overall disparities in WERs if at all.

\subsection{Mixed-Effects Poisson Regression}
Block-structured evaluation data arises naturally in any real-world speech recognition applications. In particular, utterances from the same speaker could share common correlated features (e.g.~accent of speaker), and thus analyses that assume independence of these observations will be inappropriate. The use of random effect \cite{baltagi2008econometric, faraway2016extending} is one usual and convenient way to model such structure.

Suppose we want to investigate the effect of race on speech recognition accuracy across a sample of speakers. Typically, we would treat the racial effect as fixed in the regression. On the other hand, it makes most sense to treat the speaker effect as random. It is reasonable to consider these speakers as being randomly selected from a larger collection of speakers whose characteristics we would like to estimate. We are not particularly interested in these specific speakers, but in the whole population. Generally, blocking factors can often be viewed as random effects.

A mixed-effects Poisson regression is a model containing both fixed effect and random effect. Regarding the fairness measurement of speech recognition accuracy among different subgroups of the factor $f(\cdot)$, we describe the model in detail as follows:
\begin{align}
\label{model_v3_random}
r_i & \overset{\text{i.i.d.}}{\sim} \mathcal{N}(0, \sigma^2) \\
\label{model_v3_iid}
C_{ij} \,| \, \lambda_{ij} & \overset{\text{i.i.d.}}{\sim} Poisson(\lambda_{ij}) \\
\label{model_v3_formula}
\log(\lambda_{ij})& = \log(N_{ij}) + \mu_{f(i)} + r_i + \theta^{T} x_{ij}
\end{align}
where the utterance-level index of subscription notation $ij$ represents the $j$th utterance from the $i$th speaker, $r_i$ denotes the speaker-level random effect that is independently sampled from a Gaussian distribution with mean 0 and variance $\sigma^2$ which is learnable. Note that any $C_{ij}$ and $C_{ij'}$ are no longer independent for $j \neq j'$ since they are observed from the same speaker $i$, while any $C_{i\cdot}$ and $C_{i'\cdot}$ are still independent for $i \neq i'$ since they are observed from different speakers. Also, we use $\mu_{f(i)}$ to denote the fixed effect for the factor $f(\cdot)$ of primary interest, since typically it is at speaker level.

This mixed-effects model can be fitted via maximum likelihood and the expression for its likelihood is an integral over the random effect, which must be approximated, for example, via adaptive Gauss-Hermite quadrature \cite{abramowitz1964handbook}. Again, LRT can be performed to calculate the $p$-value of the factor $f(\cdot)$ on measured WER results. In practice, it would be particularly interesting to extract the conditional modes of the speaker-level random effect for subsequent analysis and assumption verification.

\section{Simulation Experiments}
\label{simulation}
In this section, we conduct simulation experiments to show that the proposed mixed-effects Poisson regression could properly address the problems of confounding factor and speaker effect in ASR fairness measurements.

\subsection{Experiment on Confounding Factor}
We generate synthetic data to investigate the effect of confounding factor on ASR fairness measurements over \emph{case} group and \emph{control} group, defined by some primary factor of interest.

Under the scenario that recognition errors from different utterances are independent from each other, the number of errors on the $s$th utterance is randomly sampled from a Poisson distribution with the mean parameter written as
\begin{align}
    \lambda_s=N_s\cdot\exp(\mu_{f(s)} + \theta_s\cdot\emph{Bernoulli}(p_{f(s)}))
\end{align}
where $f(s)\in\{{\text{case}}, \text{{control}}\}$ indicates which group the utterance comes from, $N_s$ denotes the number of words in the reference, $\mu_{f(s)}$ refers to the group effect, $\theta_s$ represents the effect of confounding factor, and $p_{f(s)}$ is the mean parameter of a Bernoulli distribution which controls the frequency that the confounding effect is present in the corresponding group.

In our experiment, we set $N_s=10$, $\mu_{f(s)}=\log(0.05)$, $\theta_s=0.1$ for every $s$, and $p_{f(s)}$ is varied at 50\%, 60\%, 70\%, 90\% for the case group, and 50\%, 40\%, 30\%, 10\% for the control group, respectively. For each of case or control group, we generate 5,000 utterances independently.

Here, we would like to evaluate the ratio of WERs between the case and control groups, and in particular, conduct statistical testing to determine whether significant difference on WERs exists between the two groups. Based on our setup, the ground truth WER ratio is 1.0, that is, in theory there is no WER difference between the two groups. Notice that the presence of confounding factor could introduce nuisance and mislead the results since it raises up the mean number of errors by $\exp(0.1)-1\approx11\%$ at utterance level.

In this study, we consider the \textbf{baseline} measurement method as the one that computes the ratio of empirical WER of case group over the one of control group. The \emph{bootstrap} method \cite{efron1994introduction, efron2003second} is applied to compute the 95\% confidence interval (CI) of the ratio. Then if the CI does not cover the point of 1.0, we claim the WER gap between the two groups is statistically significant. Regarding \textbf{model-based} approach, we fit a Poisson regression according to (\ref{iid})~(\ref{model_v2}) which linearly incorporates the confounding factor, and then compute the 95\% CI associated with the group effect ratio.

For each approach, we repeat the simulation for 1,000 times and compare the average of estimated WER ratios as well as the false positive rate, that is, the frequency of times that the statistical significance on WER ratio is falsely claimed by the method. Strictly speaking, a 95\% CI means that if we were able to have 100 different datasets from the same distribution of the original data and compute a 95\% CI based on each of these datasets, then approximately 95 of these 100 CIs will contain the true value of the statistic of interest \cite{neyman1937x, stuart1963advanced, cox1979theoretical}. Thus in theory, we expect 5\% false positive rate if the method works correctly and generates valid CIs.

The result is shown in Table~\ref{tab:simulation1}, where we can see that the mean ratio and false positive rate of the baseline method increase dramatically when the confounding rate differs more and more between case and control groups. This is expected since the baseline method does not take into account the information of confounding factor which does harm to the inference. For model-based approach, we observe the mean ratios are around 1.0 and the false positive rates are around 5\% for all the setups, which demonstrates that it can successfully address the confounding effect and result in valid estimates of WER ratios and corresponding CIs.

\begin{table}
  \centering
  \caption{Simulation result on confounding factor experiment with various confounding rates $p_{\text{case}}$ and $p_{\text{control}}$ across groups.}
  \label{tab:simulation1}
  \resizebox{\columnwidth}{!}{%
  \begin{tabular}{cc|cr|cc}
    \toprule
    \multicolumn{2}{c|}{\emph{\;\;\;Confounding Rate\;\;}} & \multicolumn{2}{c}{\bf{Baseline}} & \multicolumn{2}{|c}{\bf{Model-Based}} \\
    \cmidrule(r){3-4}
    \cmidrule(r){5-6}                                   
    \shortstack{\emph{within} \\ \emph{Case}} & \shortstack{\emph{within} \\ \emph{Control}} & \shortstack{\emph{Mean} \\ \emph{Ratio}} & \shortstack{\emph{\% False} \\ \emph{Positive}} & \shortstack{\emph{Mean} \\ \emph{Ratio}} & \shortstack{\% \emph{False} \\ \emph{Positive}} \\
    \midrule
    50\% & 50\% & 1.000 & 4.9\% & 1.000 & 4.7\%  \\
    60\% & 40\% & 1.021 & 12.1\% & 1.001 & 5.8\% \\
    70\% & 30\% & 1.041 & 29.8\% & 1.000 & 5.4\% \\
    90\% & 10\% & 1.084 & 83.3\% & 1.001 & 5.1\% \\
    \bottomrule
  \end{tabular}
  }
\end{table}

\subsection{Experiment on Speaker Effect}
In this experiment, we generate synthetic data to study the impact of speaker effect on ASR fairness measurements of the two groups.

For any of case or control group, assume there are $I$ distinct speakers and each speaker has equal number of utterances. For the $i$th speaker and $j$th utterance from the speaker, the number of errors is sampled from a Poisson distribution with the mean parameter written as
\begin{align}
    \lambda_{ij}=N_{ij}\cdot\exp(\mu_{f(i)} + r_i),\;r_i & \overset{\text{i.i.d.}}{\sim} \mathcal{N}(0, \sigma^2)
\end{align}
where $f(i)$ indicates which group the speaker is from, $N_{ij}$ denotes the number of words in the reference, $\mu_{f(i)}$ refers to the group effect, and $r_i$ represents the speaker effect drawn from a Gaussian distribution with mean 0 and standard deviation $\sigma$.

In our experiment, we set $N_{ij}=10$, $\mu_{f(i)}=\log(0.05)$ for every $i, j$, number of speakers $I$ is varied at 100, 500 and standard deviation $\sigma$ is varied at 0.2, 0.4. For any of case or control group, we generate 5,000 utterances.

Again, we want to evaluate the ratio of WERs and the ground truth shall be 1.0. The baseline method is the same with the one used for confounding factor experiment while for the model-based approach, we fit a mixed-effects Poisson regression according to (\ref{model_v3_random})~(\ref{model_v3_iid})~(\ref{model_v3_formula}) which treats the speaker identification as a random effect with learnable $\sigma$.

The result is shown in Table~\ref{tab:simulation2}, where we can see that the mean ratios of both methods are around 1.0. However, for the baseline method, we observe high false positive rates and in particular, the higher the standard deviation $\sigma$ or the smaller the number of speakers, the larger the false positive rate. Instead, the model-based approach always results in approximate 5\% false positive rate. This demonstrates that it can successfully deal with speaker effect and is superior than the traditional baseline method.

\begin{table}
  \centering
  \caption{Simulation result on speaker effect experiment with various numbers of speakers and values of standard deviation $\sigma$.}
  \label{tab:simulation2}
  \resizebox{\columnwidth}{!}{%
  \begin{tabular}{cc|cr|cc}
    \toprule
    \multicolumn{2}{c|}{\emph{Speaker Effect}} & \multicolumn{2}{c}{\bf{Baseline}} & \multicolumn{2}{|c}{\bf{Model-Based}} \\
    \cmidrule(r){3-4}
    \cmidrule(r){5-6}                                   
    \shortstack{\emph{Num of} \\ \emph{Speakers}} & \shortstack{\emph{Standard} \\ \emph{Deviation}} & \shortstack{\emph{Mean} \\ \emph{Ratio}} & \shortstack{\emph{\% False} \\ \emph{Positive}} & \shortstack{\emph{Mean} \\ \emph{Ratio}} & \shortstack{\% \emph{False} \\ \emph{Positive}} \\
    \midrule
    500 & 0.2 & 1.000 & 8.0\% & 1.000 & 4.8\%  \\ 
    500 & 0.4 & 1.001 & 14.9\% & 1.001 & 4.5\%  \\    
    100 & 0.2 & 1.000 & 16.6\% & 1.000 & 5.0\%  \\
    100 & 0.4 & 0.999 & 42.6\% & 0.999 & 5.2\% \\
    \bottomrule
  \end{tabular}
  }
\end{table}

\section{Real Data Experiments}
\label{real}
In this section, we apply the proposed mixed-effects Poisson regression on real-world speech datasets for fairness investigation.

\subsection{Datasets and Setup}
We consider the following two ASR datasets in the experiments:
\begin{itemize}
\item \emph{LibriSpeech} \cite{panayotov2015librispeech}. A widely used voice dataset which consists of 960 hours transcribed training utterances. 
The evaluation dataset has the splits of \emph{Test-Clean} from 40 speakers and \emph{Test-Other} from 33 speakers.
\item \emph{Voice Command}. This is a de-identified dataset collected using mobile devices through crowd-sourcing from a data supplier for ASR. No personally identifiable information (PII) is contained in this dataset. The participants are instructed to say voice commands on the topics of calling friends, playing music, etc. It consists of 2,440 hours transcribed training utterances. 
The evaluation set contains around 18K utterances from 95 speakers.
\end{itemize}
Table~\ref{tab:data} shows details of the two evaluation datasets on number of utterances and number of speakers.

The ASR system in this investigation is an RNN-T model with Emformer encoder \cite{emformer2021streaming}, LSTM predictor, and a joiner, having approximately 80 million parameters in total. For each of \emph{LibriSpeech} or \emph{Voice Command} data, the ASR model is trained from scratch using the corresponding training utterances.

\subsection{Evaluation Results}
For the \emph{LibriSpeech} data, we study the ASR fairness on \emph{gender}, that is, we would like to test whether there exists statistical significance on the WER ratio between male speakers and female speakers.

The \textbf{baseline} approach, which is widely used in practice, computes the ratio of empirical WER from male speakers group over the empirical WER from female speakers group. The bootstrap method is applied to compute the 95\% CI of the ratio. For \textbf{model-based} approach, we fit a mixed-effects Poisson regression based on (\ref{model_v3_random})~(\ref{model_v3_iid})~(\ref{model_v3_formula}) with gender as the fixed effect and speaker label as a random effect.

\begin{table}
 \caption{Summary of LibriSpeech and Voice Command evaluation datasets in the experiments of real-world data analysis.}
  \centering
  \resizebox{\columnwidth}{!}{%
  \begin{tabular}{l|c|c|c}
    \toprule
    & \multicolumn{3}{|c}{\bf{Evaluation Dataset}} \\
    \cmidrule(r){2-4}    
    \emph{Feature}& \shortstack{\emph{LibriSpeech} \\ \emph{Test-Clean}} & \shortstack{\emph{LibriSpeech} \\ \emph{Test-Other}} & \shortstack{\emph{Voice} \\ \emph{Command}} \\
    \midrule
    \emph{\# of Utterances} & 2,620 & 2,939 & 17,783\\
    \emph{\# of Speakers} & 40 & 33 & 95 \\
    \emph{\# of Male Speakers} & 20 & 16 & 41 \\
    \bottomrule
  \end{tabular}
  }
  \label{tab:data}
\end{table}

Result is shown in Table~\ref{tab:real1}. We can see that the baseline method leads to statistically significance claims on both \emph{Test-Clean} and \emph{Test-Other} sets, and interestingly, their conclusions are actually opposite. Specifically, on \emph{Test-Clean} split of evaluation dataset, the baseline method shows that male speakers group has significant lower WER compared to female speakers group, while on \emph{Test-Other} split, male speakers group has significant higher WER compared to the group of female speakers. On the other hand, the model-based approach does not claim any significant results on both splits. This makes sense since numbers of speakers in both splits are quite small, which results in high variance estimation that does not lead to statistically significant results. Thus utterances from more speakers are needed to reduce the standard errors and draw a more sound conclusion.

To further trace the source of WER gap, Table~\ref{tab:real2} shows the result of mixed-effect Poisson regression with sentence embedding of the true reference text as extra explanatory variables. We use pre-trained fastText word embeddings \cite{bojanowski2017enriching} with 300 dimensions and take their average to obtain the representation at sentence level. From the result, after excluding the effect from grammatical, lexical, or semantic characteristics, the WER gap between the two groups become smaller. Although it is not statistically significant, acoustic characteristics appear to contribute to the WER disparity on \emph{Test-Other}.

\begin{table}
  \centering
  \caption{Real-world analysis result on LibriSpeech dataset.}
  \label{tab:real1}
  \begin{tabular}{c|cr|cc}
    \toprule
    & \multicolumn{2}{c}{\bf{Baseline}} & \multicolumn{2}{|c}{\bf{Model-Based}} \\
    \cmidrule(r){2-3}
    \cmidrule(r){4-5}
    \shortstack{\emph{LibriSpeech} \\ \emph{Dataset}} & \shortstack{\emph{WER} \\ \emph{Ratio}} & \shortstack{\emph{Confidence} \\ \emph{Interval}} & \shortstack{\emph{WER} \\ \emph{Ratio}} & \shortstack{\emph{Confidence} \\ \emph{Interval}} \\
    \midrule
    \emph{Test-Clean} & 0.86 & (0.76, 0.97) & 0.88 & (0.67, 1.14) \\ 
    \emph{Test-Other} & 1.34 & (1.23, 1.46) & 1.28 & (0.93, 1.76) \\
    \bottomrule
  \end{tabular}
\end{table}

\begin{table}
  \centering
  \caption{Real-world analysis result on LibriSpeech dataset with sentence embedding as explanatory variables.}
  \label{tab:real2}
  \begin{tabular}{c|cr}
    \toprule
    & \multicolumn{2}{c}{\bf{Model-Based (Embed)}} \\
    \cmidrule(r){2-3}
    \shortstack{\emph{LibriSpeech} \\ \emph{Dataset}} & \shortstack{\emph{WER} \\ \emph{Ratio}} & \shortstack{\emph{Confidence} \\ \emph{Interval}} \\
    \midrule
    \emph{Test-Clean} & 1.01 & (0.76, 1.33) \\ 
    \emph{Test-Other} & 1.19 & (0.87, 1.64) \\
    \bottomrule
  \end{tabular}
  \vspace{-0.125cm}
\end{table}

\begin{table}[ht]
  \vspace{-0.125cm}
  \centering
  \caption{Real-world analysis result on Voice Command dataset.}
  \label{tab:real3}
  \begin{tabular}{c|cr|cc}
    \toprule
    & \multicolumn{2}{c}{\bf{Baseline}} & \multicolumn{2}{|c}{\bf{Model-Based}} \\
    \cmidrule(r){2-3}
    \cmidrule(r){4-5}
    \shortstack{\emph{Voice Command} \\ \emph{Dataset}} & \shortstack{\emph{WER} \\ \emph{Ratio}} & \shortstack{\emph{Confidence} \\ \emph{Interval}} & \shortstack{\emph{WER} \\ \emph{Ratio}} & \shortstack{\emph{Confidence} \\ \emph{Interval}} \\
    \midrule
    \emph{Test} & 1.08 & (0.99, 1.20) & 1.15 & (0.78, 1.72)\\ 
    \bottomrule
  \end{tabular}
\end{table}

We also investigate ASR fairness on gender for \emph{Voice Command} dataset. The baseline and model-based methods are the same with the ones applied for \emph{LibriSpeech}. Result is shown in Table~\ref{tab:real3}. The baseline method does not claim that the WER on male speakers group is statistically significantly higher than the WER of female speakers group, but it is very close. The model-based method clearly does not lead to significant result, due to the relatively small number of speakers in each group.

\section{Conclusions}
\label{conclusion}
In this paper, we introduce mixed-effects Poisson regression to better measure and interpret any WER difference among subgroups of interest. The presented method is very flexible to use and can effectively address the open problems of how to control the nuisance factors, how to handle unobserved heterogeneity across speakers, and how to trace the source of any WER gap among different subgroups.


\bibliographystyle{IEEEbib}
\bibliography{strings,refs}

\end{document}